\documentclass{article}

\PassOptionsToPackage{numbers, sort}{natbib}

\usepackage[final]{neurips_2025}

\usepackage[utf8]{inputenc} 
\usepackage[T1]{fontenc}    
\usepackage{hyperref}       
\usepackage{url}            
\usepackage{booktabs}       
\usepackage{amsfonts}       
\usepackage{nicefrac}       
\usepackage{microtype}      
\usepackage{xcolor}         

\usepackage{minted}
\usepackage{listings}

\usepackage{amsmath}
\usepackage{amssymb}
\usepackage{mathtools}
\usepackage{amsthm}
\usepackage{float}
\usepackage[capitalize,noabbrev]{cleveref}
\usepackage{multirow}
\usepackage{microtype}
\usepackage{graphicx}
\usepackage{mdframed}
\usepackage{subfigure}
\usepackage{makecell}
\usepackage{booktabs} 
\usepackage{enumitem}
\usepackage{wrapfig}
\usepackage{caption}
\usepackage[normalem]{ulem}
\useunder{\uline}{\ul}{}
\usepackage{xurl}
\usepackage{longtable}
\usepackage[table,xcdraw]{xcolor}

\usepackage{lipsum}
\usepackage[page]{appendix}

\renewcommand{\appendixtocname}{List of appendices}

\makeatletter
\let\oldappendix\appendices

\renewcommand{\appendices}{%
  \clearpage
  \renewcommand{\thesection}{\Roman{section}}
  \let\tf@toc\tf@app
  \addtocontents{app}{\protect\setcounter{tocdepth}{2}}
  \immediate\write\@auxout{%
    \string\let\string\tf@toc\string\tf@app^^J
  }
  \oldappendix
}%

\newcommand{\listofappendices}{%
  \begingroup
  \renewcommand{\contentsname}{\appendixtocname}
  \let\@oldstarttoc\@starttoc
  \def\@starttoc##1{\@oldstarttoc{app}}
  \tableofcontents
  \endgroup
}

\makeatother

\usepackage[most]{tcolorbox}
\newtcolorbox{promptbox}{
  enhanced,
  breakable, 
  colback=gray!10,
  colframe=gray!70,
  boxrule=0.5mm,
  arc=0pt,
  top=10pt,
  bottom=10pt,
  left=10pt,
  right=10pt,
}

\hypersetup{
    colorlinks=true,
    linkcolor=red,
    citecolor=blue,
    filecolor=magenta,      
    urlcolor=blue,
linktocpage}

\title{What Affects the Effective Depth of Large Language Models?}

\author{%
  Yi Hu\textsuperscript{1} \quad
  Cai Zhou\textsuperscript{2} \quad
  Muhan Zhang\textsuperscript{1,\dag} \\
  \textsuperscript{1}Institute for Artificial Intelligence, Peking University \\
  \textsuperscript{2}Department of Computer Science, Massachusetts Institute of Technology \\
}

\begin{document}

\maketitle
\begin{abstract}
\label{abstract}
The scaling of large language models (LLMs) emphasizes increasing depth, yet performance gains diminish with added layers. Prior work introduces the concept of ``effective depth'', arguing that deeper models fail to fully utilize their layers for meaningful computation. Building on this, we systematically study how effective depth varies with model scale, training type, and task difficulty.
First, we analyze the model behavior of Qwen-2.5 family (1.5B–32B) and find that while the number of effective layers grows with model size, the effective depth ratio remains stable. Besides, comparisons between base and corresponding long-CoT models show no increase in effective depth, suggesting that improved reasoning stems from longer context rather than deeper per-token computation. Furthermore, evaluations across tasks of varying difficulty indicate that models do not dynamically use more layers for harder problems.
Our results suggest that current LLMs underuse available depth across scales, training paradigms and tasks of varying difficulties, pointing out research opportunities on increasing the layer utilization rate of LLMs, model pruning, and early exiting. Our code is released at \url{https://github.com/AheadOFpotato/what_affects_effective_depth}.

\end{abstract}
\section{Introduction}
\label{sec:introduction}

The scaling of large language models (LLMs)~\citep{yang2025qwen3,qwen2025qwen25technicalreport,grattafiori2024llama3herdmodels,deepseekai2024deepseekv3technicalreport,achiam2023gpt} has consistently emphasized increased depth, with empirical evidence suggesting that model performance improves with additional layers---despite diminishing returns. As pointed out by~\citet{eff_depth}, this trend raises a fundamental question: are these models truly leveraging their depth to perform more complex, hierarchical computations, or are they merely distributing similar computational operations over a greater number of layers? 

\citet{eff_depth} reveals a striking under-utilization of depth: layers in the second half are simply refining existing representations rather than contributing to novel feature composition or conducting deeper reasoning. The study introduces the concept of ``effective depth'' and suggests that inefficient depth utilization may be a fundamental cause of diminishing scaling returns. Building directly upon this foundation, our work seeks to systematically investigate the factors that influence this effective depth. We aim to achieve a more comprehensive understanding of how depth utilization behaves across model scale, specialized training, and task difficulty. Our findings are as follows:
\begin{enumerate}[
  topsep=0pt,      
  partopsep=0pt,   
  itemsep=0pt,     
  parsep=0pt,      
  leftmargin=*,    
]
    \item \textbf{Regarding model size.} Following the methodologies established in prior work, we first analyze the Qwen-2.5 model family (from 1.5B to 32B)~\citep{qwen2025qwen25technicalreport} using a suite of techniques including residual cosine similarity, logit lens, layer effects on future computations, residual erasure and integrated gradients~\citep{eff_depth,logitlens}. Our results confirm the core phenomenon: there exists a phase transition where early layers drive feature composition and later layers engage in minor refinements. Furthermore, while the absolute number of these ``effective'' layers increases with model size, the ratio of effective depth to total depth remains stable. This aligns with the conclusions of~\citet{eff_depth} that larger models do not fundamentally alter their computational strategy; they simply replicate the same utilization pattern over a larger number of layers, rather than using the extra depth to invent new types of computation. This finding provides a nuanced explanation for diminishing returns—wider models gain new capabilities, while deeper models primarily gain precision.
    \item \textbf{Regarding long-CoT models.} Given that long-CoT models have demonstrated exceptional performance in complex reasoning tasks~\citep{deepseekai2025deepseekr1incentivizingreasoningcapability,openaio1}, a natural hypothesis is that they might achieve this by more effectively exploiting their depth for ``deeper'' reasoning in each forward pass. To test this, we compare the effective depth of base and instruct models in the Qwen-2.5 model family~\citep{qwen2025qwen25technicalreport} against their corresponding DeepSeek-R1-distill counterparts~\citep{deepseekai2025deepseekr1incentivizingreasoningcapability}. Surprisingly, our analysis reveals no significant increase in effective depth. The enhanced reasoning performance appears not to be driven by a fundamental change in how the model utilizes its layers during each forward pass. Instead, the gains are likely attributable to the model's optimized ability to reason over longer sequences, not to deeper computation within a single token's forward process.
    \item \textbf{Regarding task difficulty.} We further probe whether models dynamically allocate their depth based on computational demand. One might expect harder problems to require and therefore activate deeper layers. We evaluate models on a difficulty spectrum from HellaSwag (natural language understanding)~\citep{hellaswag} to GSM8K (grade school math)~\citep{gsm8k} to AIME24 (high school math contests)~\citep{aime}. Counter to intuition, the effective depth remains largely consistent across all tasks. The model does not appear to leverage significantly more of its depth for harder problems.
\end{enumerate}

In summary, modern LLMs, across scales, specialized training regimes and task difficulties, fail to fully exploit their available depth for composing novel, high-level features.

\section{Preliminary}
We mainly focus on the Qwen-2.5 model family~\citep{qwen2025qwen25technicalreport} (including base models and instruct models), and their corresponding DeepSeek-R1-Distill versions~\citep{deepseekai2025deepseekr1incentivizingreasoningcapability}. They are all pre-norm Transformers~\citep{xiong2020layer,vaswani2017attention} and the forward process of a layer $l$ is as follows:
\begin{align}
\boldsymbol{a}_l&=\mathrm{SelfAttention}_l(\mathrm{RMSNorm}(\boldsymbol{h}_l))\\\label{eq1}
\hat{\boldsymbol{h}}_l&=\boldsymbol{h}_l+\boldsymbol{a}_l\\\label{eq2}
\boldsymbol{m}_{l}&=\mathrm{MLP}_l(\mathrm{RMSNorm}(\hat{\boldsymbol{h}}_l))\\\label{eq3}
\boldsymbol{h}_{l+1}&=\hat{\boldsymbol{h}}_l+\boldsymbol{m}_l
\end{align}

Here, $\boldsymbol{h}_l\in\mathbb{R}^{n_{\mathrm{context}}\times d_{\mathrm{model}}}$ is the residual stream~\citep{elhage2021mathematical}, $\boldsymbol{a}_l, \boldsymbol{m}_l$ are the outputs of the SelfAttention layers and MLP layers, which are directly added back to the residual stream. $n_{\mathrm{context}}$ is the length of the input sequence, and $d_{\mathrm{model}}$ is the dimension of the hidden states of the model. RMSNorm~\citep{zhang2019root} is adopted in the Qwen-2.5 model family to replace traditional layer normalization~\citep{xiong2020layer}. Following~\citet{eff_depth}, we denote $\mathrm{SelfAttention}_l(\cdot)$ and $\mathrm{MLP}_l(\cdot)$ as ``sublayers''.

The residual stream starts with $\boldsymbol{h}_0 = \mathrm{Embedding}(x)$, where $x\in\mathbb{N}^{n_{\mathrm{context}}}$ is the sequence of \texttt{token\_id}s. Then the final results of residual stream goes through the output layer and results in the output probability distribution over vocabulary: $\boldsymbol{y} = \mathrm{softmax}(\mathrm{RMSNorm}(\boldsymbol{h}_{L})\boldsymbol{W}^{out})$, where $\boldsymbol{y}\in \mathbb{R}^{n_{\mathrm{context}}\times |V|}$, $\boldsymbol{W}^{out}\in\mathbb{R}^{d_{\mathrm{model}}\times |V|}$, $L$ is the number of layers in the model, $V$ is the vocabulary.

\section{Methods}

\citet{eff_depth} proposes a suite of methods to qualitatively probe the effective depth. We introduce and extend the methods to qualitatively assess effective depth across different models and datasets:

\textbf{Residual cosine similarity.} Residual cosine similarity measures how each layer or sublayer interacts with the residual stream. For a given layer \(l\), we compute the cosine similarity between its contribution (the output of either $\mathrm{SelfAttention}$ \(\boldsymbol{a}_l\), $\mathrm{MLP}$ \(\boldsymbol{m}_l\), or their sum) and the resulting residual state \(\boldsymbol{h}_l\). Formally, the similarities are defined as \(\mathrm{cosim}(\boldsymbol{a}_l + \boldsymbol{m}_l, \boldsymbol{h}_l)\) for the full layer, \(\mathrm{cosim}(\boldsymbol{a}_l, \boldsymbol{h}_l)\) for self-attention, and \(\mathrm{cosim}(\boldsymbol{m}_l, \boldsymbol{h}_l + \boldsymbol{a}_l)\) for the MLP. The intuition is that a cosine similarity near zero suggests the module writes a new, orthogonal feature into the residual stream; negative values indicate feature erasure; and positive values signify the amplification of an existing feature.

\textbf{Logit Lens.} Logit lens evaluates how early the model’s output distribution begins to stabilize. We decode the hidden state \(\boldsymbol{h}_l\) using the model’s output projection and compute the KL divergence between this early distribution and the model’s final distribution. Additionally, we measure the overlap between the top-5 tokens from this intermediate distribution and from the final distribution.

\textbf{Layer effects on future computation.} Here we probe the influence of skipping a layer on subsequent computations. For a given prompt, we first run a forward pass to record the residual states \(h_l\). We then intervene by skipping a specific layer \(s\) for all token positions \(t \leq t_s\) (where \(t_s\) is a sampled position within the sequence), effectively setting \(\bar{h}_{s+1} := \bar{h}_s\) for those tokens. The effect of this intervention is measured on the subsequent tokens (\(t > t_s\)) by computing the relative change in the contribution of a later layer \(l > s\): \(\| (h_{l+1} - h_l) - (\bar{h}_{l+1} - \bar{h}_l) \|_2 / \| h_{l+1} - h_l \|_2\). The maximum value of this metric across multiple prompts and sequence positions is taken. We also compare the final output probabilities directly via \(\| y - \bar{y} \|_2\).

\textbf{Residual erasure.} Residual erasure identifies until which layer information from a specific token remains relevant for the final prediction. For a token at position \(t\) and layer \(l\), we intervene by replacing its residual vector \(h_{l+1}[t]\) with an uninformative baseline—the average residual vector at layer \(l\) computed over a dataset (GSM8K here), while leaving all other tokens unchanged. The effect is quantified as the maximum change in prediction norm (\( \| y - \bar{y} \|_2 \)) among all answer tokens. 

\textbf{Integrated gradients.} The metric attributes the model’s prediction on the answer tokens to contributions from each layer. We compute the gradients of the output logits for all answer tokens with respect to the activation at each layer and each token position.

\label{method:quantitative}
Beyond these qualitative probes, we introduce two quantitative measures to compare effective depth across models and datasets. For \textit{residual cosine similarity}, we average the similarity scores across layers, $\mathrm{MLP}$s, and $\mathrm{SelfAttention}$ modules, and identify the effective depth as the point where the averaged similarity transitions from negative to positive. For the \textit{logit lens}, we use two metrics: we define the effective depth as the layer where the KL divergence from the final output drops below half of its maximum observed value, and alternatively, as the layer where the top-5 token overlap with the final output first exceeds 0.3.
\begin{table}[]
\caption{Effective depth (ED) and effective depth ratio ($\text{ratio} = \frac{\text{ED}+1}{\text{L}}$) across base, instruct, and long-CoT models of different sizes (1.5B to 32B parameters) and on datasets with varying difficulty.}
\centering
\resizebox{\textwidth}{!}{
\begin{tabular}{l|c|c|c|c|c|c|c|c|c|c|c|c|c|c|c|c|c|c}
\toprule
\textbf{} & \multicolumn{6}{c|}{\textbf{Cosine Similarity}}& \multicolumn{6}{c|}{\textbf{Logit Lens KL}}    & \multicolumn{6}{c}{\textbf{Logit Lens Overlap}}                   \\
\textbf{} & \multicolumn{2}{c}{HellaSwag}     & \multicolumn{2}{c}{GSM8K}         & \multicolumn{2}{c|}{AIME24}        & \multicolumn{2}{c}{HellaSwag}     & \multicolumn{2}{c}{GSM8K}         & \multicolumn{2}{c|}{AIME24}        & \multicolumn{2}{c}{HellaSwag}     & \multicolumn{2}{c}{GSM8K}         & \multicolumn{2}{c}{AIME24}        \\
\textbf{} & \multicolumn{1}{c}{ED} & \multicolumn{1}{c}{ratio}    & \multicolumn{1}{c}{ED} & \multicolumn{1}{c}{ratio}    & \multicolumn{1}{c}{ED} & \multicolumn{1}{c|}{ratio}    & \multicolumn{1}{c}{ED} & \multicolumn{1}{c}{ratio}    & \multicolumn{1}{c}{ED} & \multicolumn{1}{c}{ratio}    & \multicolumn{1}{c}{ED} & \multicolumn{1}{c|}{ratio}    & \multicolumn{1}{c}{ED} & \multicolumn{1}{c}{ratio}    & \multicolumn{1}{c}{ED} & \multicolumn{1}{c}{ratio}    & \multicolumn{1}{c}{ED} & \multicolumn{1}{c}{ratio}    \\
\midrule
DS-R1-Qwen-1.5B & 17 & \cellcolor[HTML]{FDC67D}0.64 & 16 & \cellcolor[HTML]{FCC17C}0.61 & 17 & \cellcolor[HTML]{FDC67D}0.64 & 20 & \cellcolor[HTML]{FED880}0.75 & 1  & \cellcolor[HTML]{F8696B}0.07 & 24 & \cellcolor[HTML]{D5DF82}0.89 & 23 & \cellcolor[HTML]{FFEB84}0.86 & 23 & \cellcolor[HTML]{FFEB84}0.86 & 24 & \cellcolor[HTML]{D5DF82}0.89 \\
Qwen2.5-1.5B-Instruct         & 16 & \cellcolor[HTML]{FCC17C}0.61 & 20 & \cellcolor[HTML]{FED880}0.75 & 19 & \cellcolor[HTML]{FDD27F}0.71 & 21 & \cellcolor[HTML]{FEDF81}0.79 & 22 & \cellcolor[HTML]{FEE482}0.82 & 23 & \cellcolor[HTML]{FFEB84}0.86 & 23 & \cellcolor[HTML]{FFEB84}0.86 & 23 & \cellcolor[HTML]{FFEB84}0.86 & 23 & \cellcolor[HTML]{FFEB84}0.86 \\
Qwen2.5-Math-1.5B             & 16 & \cellcolor[HTML]{FCC17C}0.61 & 16 & \cellcolor[HTML]{FCC17C}0.61 & 16 & \cellcolor[HTML]{FCC17C}0.61 & 20 & \cellcolor[HTML]{FED880}0.75 & 22 & \cellcolor[HTML]{FEE482}0.82 & 23 & \cellcolor[HTML]{FFEB84}0.86 & 23 & \cellcolor[HTML]{FFEB84}0.86 & 23 & \cellcolor[HTML]{FFEB84}0.86 & 23 & \cellcolor[HTML]{FFEB84}0.86 \\
\midrule
DS-R1-Qwen-7B   & 16 & \cellcolor[HTML]{FCC17C}0.61 & 16 & \cellcolor[HTML]{FCC17C}0.61 & 16 & \cellcolor[HTML]{FCC17C}0.61 & 24 & \cellcolor[HTML]{D5DF82}0.89 & 24 & \cellcolor[HTML]{D5DF82}0.89 & 24 & \cellcolor[HTML]{D5DF82}0.89 & 25 & \cellcolor[HTML]{9CCF7F}0.93 & 25 & \cellcolor[HTML]{9CCF7F}0.93 & 24 & \cellcolor[HTML]{D5DF82}0.89 \\
Qwen2.5-7B-Instruct           & 17 & \cellcolor[HTML]{FDC67D}0.64 & 20 & \cellcolor[HTML]{FED880}0.75 & 18 & \cellcolor[HTML]{FDCD7E}0.68 & 25 & \cellcolor[HTML]{9CCF7F}0.93 & 25 & \cellcolor[HTML]{9CCF7F}0.93 & 25 & \cellcolor[HTML]{9CCF7F}0.93 & 26 & \cellcolor[HTML]{72C37C}0.96 & 26 & \cellcolor[HTML]{72C37C}0.96 & 26 & \cellcolor[HTML]{72C37C}0.96 \\
Qwen2.5-Math-7B               & 16 & \cellcolor[HTML]{FCC17C}0.61 & 11 & \cellcolor[HTML]{FBA476}0.43 & 16 & \cellcolor[HTML]{FCC17C}0.61 & 23 & \cellcolor[HTML]{FFEB84}0.86 & 23 & \cellcolor[HTML]{FFEB84}0.86 & 23 & \cellcolor[HTML]{FFEB84}0.86 & 24 & \cellcolor[HTML]{D5DF82}0.89 & 24 & \cellcolor[HTML]{D5DF82}0.89 & 24 & \cellcolor[HTML]{D5DF82}0.89 \\
\midrule
DS-R1-Qwen-14B  & 26 & \cellcolor[HTML]{FCB97A}0.56 & 30 & \cellcolor[HTML]{FDC87D}0.65 & 30 & \cellcolor[HTML]{FDC87D}0.65 & 40 & \cellcolor[HTML]{FEE983}0.85 & 39 & \cellcolor[HTML]{FEE683}0.83 & 41 & \cellcolor[HTML]{E3E383}0.88 & 44 & \cellcolor[HTML]{8ECB7E}0.94 & 44 & \cellcolor[HTML]{8ECB7E}0.94 & 44 & \cellcolor[HTML]{8ECB7E}0.94 \\
Qwen2.5-14B-Instruct          & 27 & \cellcolor[HTML]{FCBC7B}0.58 & 32 & \cellcolor[HTML]{FDCF7E}0.69 & 30 & \cellcolor[HTML]{FDC87D}0.65 & 40 & \cellcolor[HTML]{FEE983}0.85 & 41 & \cellcolor[HTML]{E3E383}0.88 & 42 & \cellcolor[HTML]{C7DB81}0.90  & 45 & \cellcolor[HTML]{72C37C}0.96 & 45 & \cellcolor[HTML]{72C37C}0.96 & 45 & \cellcolor[HTML]{72C37C}0.96 \\
Qwen2.5-14B                   & 27 & \cellcolor[HTML]{FCBC7B}0.58 & 30 & \cellcolor[HTML]{FDC87D}0.65 & 30 & \cellcolor[HTML]{FDC87D}0.65 & 40 & \cellcolor[HTML]{FEE983}0.85 & 40 & \cellcolor[HTML]{FEE983}0.85 & 42 & \cellcolor[HTML]{C7DB81}0.90  & 45 & \cellcolor[HTML]{72C37C}0.96 & 45 & \cellcolor[HTML]{72C37C}0.96 & 45 & \cellcolor[HTML]{72C37C}0.96 \\
\midrule
DS-R1-Qwen-32B  & 42 & \cellcolor[HTML]{FDCB7D}0.67 & 42 & \cellcolor[HTML]{FDCB7D}0.67 & 46 & \cellcolor[HTML]{FDD57F}0.73 & 58 & \cellcolor[HTML]{AAD380}0.92 & 55 & \cellcolor[HTML]{E3E383}0.88 & 57 & \cellcolor[HTML]{B9D780}0.91 & 61 & \cellcolor[HTML]{63BE7B}0.97 & 58 & \cellcolor[HTML]{AAD380}0.92 & 58 & \cellcolor[HTML]{AAD380}0.92 \\
Qwen2.5-32B-Instruct          & 43 & \cellcolor[HTML]{FDCF7E}0.69 & 46 & \cellcolor[HTML]{FDD57F}0.73 & 43 & \cellcolor[HTML]{FDCF7E}0.69 & 60 & \cellcolor[HTML]{80C77D}0.95 & 58 & \cellcolor[HTML]{AAD380}0.92 & 58 & \cellcolor[HTML]{AAD380}0.92 & 61 & \cellcolor[HTML]{63BE7B}0.97 & 60 & \cellcolor[HTML]{80C77D}0.95 & 60 & \cellcolor[HTML]{80C77D}0.95 \\
Qwen2.5-32B                   & 43 & \cellcolor[HTML]{FDCF7E}0.69 & 46 & \cellcolor[HTML]{FDD57F}0.73 & 46 & \cellcolor[HTML]{FDD57F}0.73 & 60 & \cellcolor[HTML]{80C77D}0.95 & 57 & \cellcolor[HTML]{B9D780}0.91 & 59 & \cellcolor[HTML]{8ECB7E}0.94 & 61 & \cellcolor[HTML]{63BE7B}0.97 & 59 & \cellcolor[HTML]{8ECB7E}0.94 & 60 & \cellcolor[HTML]{80C77D}0.95 \\
\bottomrule
\end{tabular}
}
\label{tab:main}
\label{-30pt}
\end{table}

\begin{figure}
    \centering
    \includegraphics[width=\linewidth]{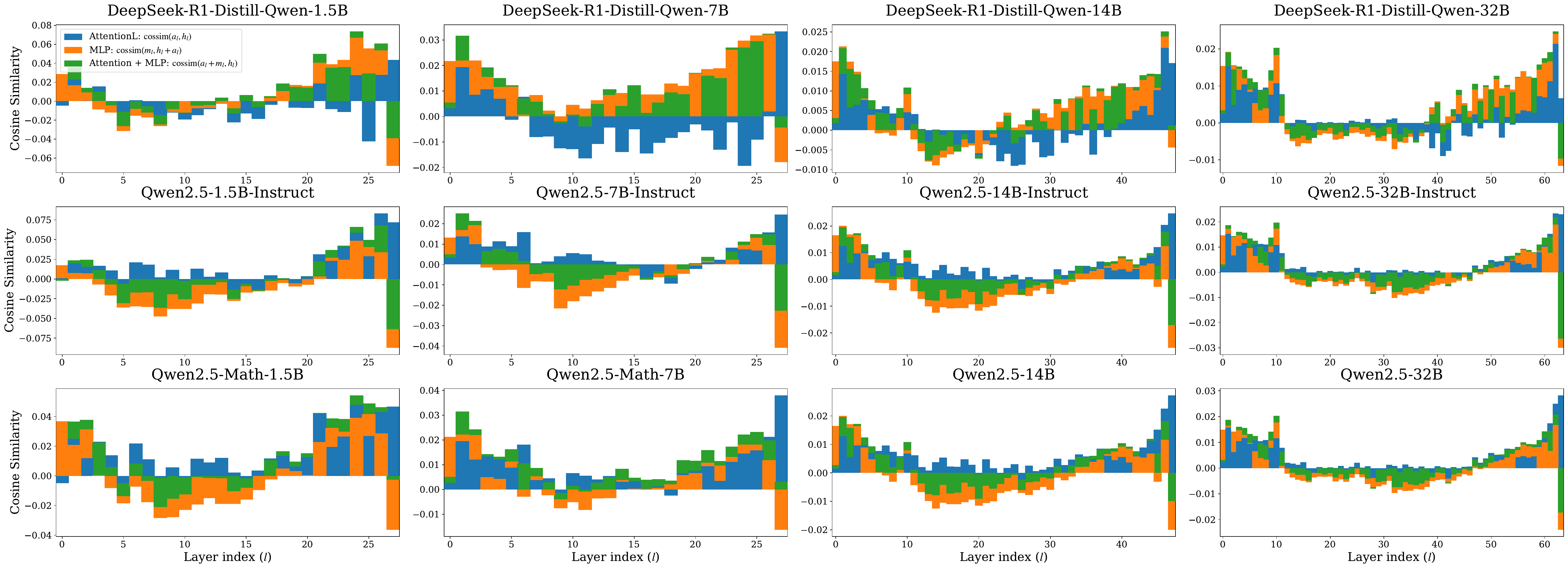}
    \caption{Cosine similarity of (sub)layer contributions and the residual evaluated on GSM8K.}
    \vspace{-20pt}
    \label{fig:gsm8k_relative_cossim_all_models}
\end{figure}

\section{Experiments}

\subsection{Does Model Size Affect Effective Depth?}

The residual cosine similarity, shown in Figure~\ref{fig:gsm8k_relative_cossim_all_models}, exhibits a consistent pattern across models: an initial positive phase, followed by a decline into negative values, and a final return to positive. The initial near-zero similarity in shallow layers suggests context integration, while the subsequent positive phase corresponds to feature refinement. The first half of the network is predominantly characterized by feature erasure (negative similarity), until a sharp phase transition occurs near the middle layers, after which the model begins strengthening existing features.

We quantify the corresponding depth of this transition in Table~\ref{tab:main} (Cosine Similarity). The results show that the effective depth ratio remains remarkably stable. This indicates that larger models contain a growing number of ``ineffective'' layers that do not contribute to feature composition.

The logit lens analysis, as shown in Figure~\ref{fig:gsm8k_logitlens}, further supports this conclusion. The KL divergence between intermediate and final predictions shows a sharp drop in the second half of the network, while the top-5 token overlap exhibits a concurrent sharp rise. Together, these indicate a transition from computation to refinement. As quantified in Table~\ref{tab:main}, the depth of this transition, measured both by KL divergence (half-max point) and overlap (exceeding 0.3), is slightly less consistent across scales than the cosine similarity metric, with a mild increasing trend in ratio for larger models.

Furthermore, the effect of skipping layers on downstream computations, illustrated in Figure~\ref{fig:gsm8k_layer_effect}, reveals that layers in the second half have substantially less influence on both later layers and final output predictions. This pattern is consistent across all model sizes, with similar decay profiles.

Finally, results from integrated gradients (Figure~\ref{fig:gsm8k_igrads_all_models}) and residual erasure (Figure~\ref{fig:gsm8k_logit_erasure_all_models}) show that the dependence of answer token predictions on earlier layers declines markedly in the second half of the network. The position of this decline remains stable relative to network depth across model sizes.

\subsection{Do Long-CoT Models Think Deeper?}

Given that long-CoT models demonstrate superior performance on complex reasoning~\citep{deepseekai2025deepseekr1incentivizingreasoningcapability, openaio1}, one might hypothesize that they achieve this by utilizing deeper computations within each forward pass. To test this, we compare the effective depth of DeepSeek-R1-Distill models~\citep{deepseekai2025deepseekr1incentivizingreasoningcapability} against their corresponding base models~\citep{qwen2025qwen25technicalreport}.
As summarized in Table~\ref{tab:main}, we find no significant difference in effective depth ratio between long-CoT and base models. This consistency is further illustrated across all probing methods: residual cosine similarity (Figure~\ref{fig:gsm8k_relative_cossim_all_models}), logit lens (Figure~\ref{fig:gsm8k_logitlens}), layer-skipping effects (Figure~\ref{fig:gsm8k_layer_effect}), integrated gradients (Figure~\ref{fig:gsm8k_igrads_all_models}), and residual erasure (Figure~\ref{fig:gsm8k_logit_erasure_all_models}). The results are consistent that long-CoT models do not exhibit a deeper utilization of the network. 

\subsection{Does Task Difficulty Affect Effective Depth?}

We next investigate whether models dynamically adjust their effective depth in response to computational demand, expecting that harder tasks might engage deeper layers. We evaluate models on three tasks of increasing difficulty: HellaSwag (natural language understanding)~\citep{hellaswag}, GSM8K (grade school math)~\citep{gsm8k}, and AIME24 (high school math contests)~\citep{aime}.
Results in Table~\ref{tab:main} show that effective depth remains largely consistent across all tasks, indicating that model depth utilization is not adaptive to problem difficulty. Additional results are provided in Appendix~\ref{app:ed_of_gsm8k_and_hellaswag}, including residual cosine similarity (Figure~\ref{fig:gsm8k_hellaswag_cosine_similarity}), effects of skipping layers on future computations (Figure~\ref{fig:gsm8k_hellaswag_layer_effects_future_compute}) and output distributions (Figure~\ref{fig:gsm8k_hellaswag_layer_effects_output_norm}), as well as logit lens KL divergence (Figure~\ref{fig:gsm8k_hellaswag_logit_lens_kl_divergence}) and token overlap (Figure~\ref{fig:gsm8k_hellaswag_logit_lens_overlap}).
\section{Conclusion}
In this work, we provide a comprehensive study of the factors that may affect effective depth in LLMs, including model scales, training strategies, and task difficulties. First, the effective depth ratio remains roughly constant with the increase of model size. Second, long-CoT models show no increase in effective depth despite their enhanced reasoning capabilities. Third, effective depth remains consistent across task difficulty, indicating no dynamic depth allocation based on computational demand.
These results demonstrate that LLMs fail to fully exploit their architectural depth.

\bibliographystyle{unsrtnat}
\bibliography{ref}

\newpage

\begin{appendices}
\section{Limitations}
\label{app:limitations}
This work follows the methodology of~\citet{eff_depth} to comprehensively analyze factors influencing effective depth. We introduce quantitative metrics based on residual cosine similarity and logit lens to compare effective depth across models and datasets. However, the proposed metrics---particularly the two variants of logit lens---remain relatively straightforward and exhibit some instability. Developing more robust and well-validated measures of effective depth is an important direction for future research.

Furthermore, while we confirm and extensively analyze the phenomenon of depth under-utilization across model scales, training strategies, and task demands, this study does not propose solutions to improve layer utilization. Our findings highlight the need for future work to explore architectural or training approaches that enable models to leverage their full depth more effectively.

\section{Model Details}
Our analysis focuses on the Qwen-2.5 model family~\citep{qwen2025qwen25technicalreport}. For base models, we use the same versions selected by~\citet{deepseekai2025deepseekr1incentivizingreasoningcapability}: Qwen2.5-Math-1.5B, Qwen2.5-Math-7B, Qwen2.5-14B, and Qwen2.5-32B. For instruction-tuned models, we use the standard instruct variants from the Qwen-2.5 family: Qwen2.5-1.5B-Instruct, Qwen2.5-7B-Instruct, Qwen2.5-14B-Instruct, and Qwen2.5-32B-Instruct. Additionally, we include the corresponding DeepSeek-R1-Distill versions derived from these base models: DeepSeek-R1-Distill-Qwen-1.5B, DeepSeek-R1-Distill-Qwen-7B, DeepSeek-R1-Distill-Qwen-14B, and DeepSeek-R1-Distill-Qwen-32B.

Models of the same size share identical architectures. The architectural details are provided in Table~\ref{tab:model}.

\begin{table}[H]
\centering
\caption{Model details.}
\begin{tabular}{c|cc}
\toprule
Models & Layers & Heads (Q/KV) \\
\midrule
1.5B   & 28     & 12 / 2       \\
7B     & 28     & 28 / 4       \\
14B    & 48     & 40 / 8       \\
32B    & 64     & 40 / 8      \\\bottomrule
\end{tabular}
\label{tab:model}
\end{table}

\section{Additional Effective Depth Results on GSM8K}

\begin{figure}[H]
    \centering
    \subfigure[Logit lens KL Divergence.]{
        \includegraphics[width=.47\linewidth]{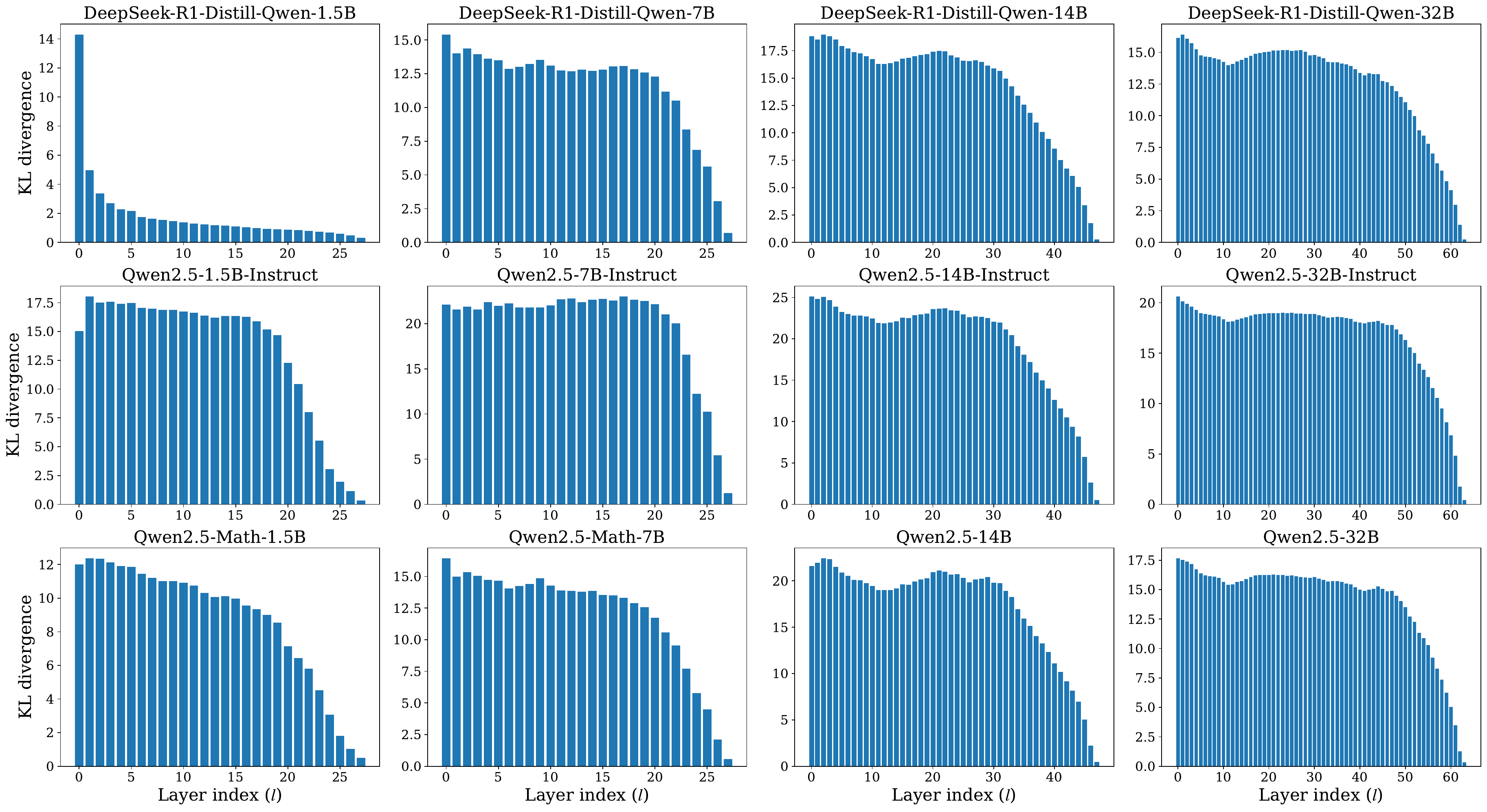}
        \label{fig:gsm8k_logit_lens_kl_all_models}
    }
    \subfigure[Logit lens Overlap.]{
        \includegraphics[width=.47\linewidth]{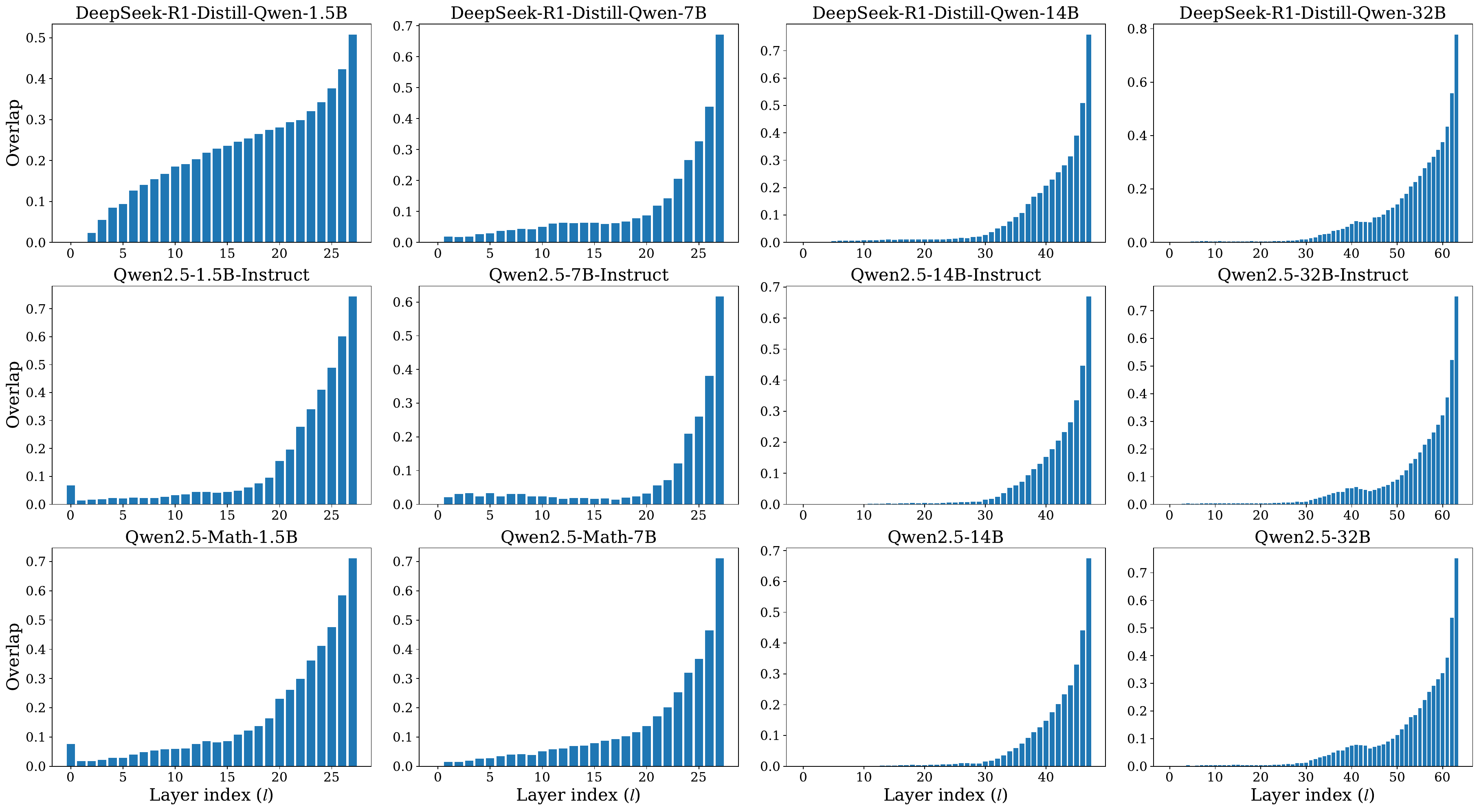}
        \label{fig:gsm8k_logit_lens_overlap_all_models}
    }
    \caption{Logit lens Results on GSM8K.}
    \label{fig:gsm8k_logitlens}
\end{figure}

\begin{figure}[H]
    \centering
    \subfigure[Effects of skipping a layer on later layers' contribution.]{
        \includegraphics[width=.4\linewidth]{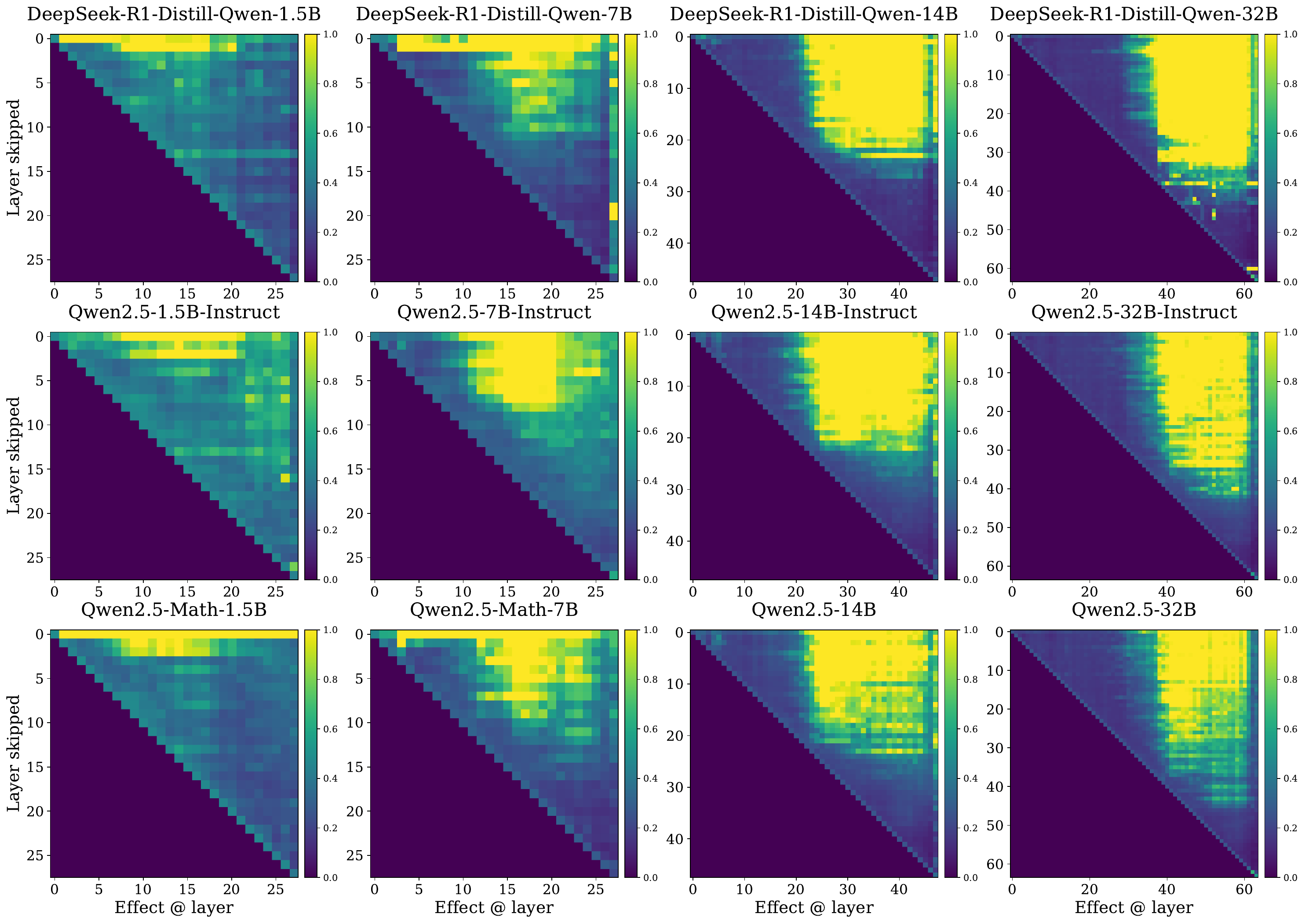}
        \label{fig:gsm8k_layer_effects_all_models}
    }
    \hspace{3pt}
    \subfigure[Effects of skipping a layer on output norm.]{
        \includegraphics[width=.52\linewidth]{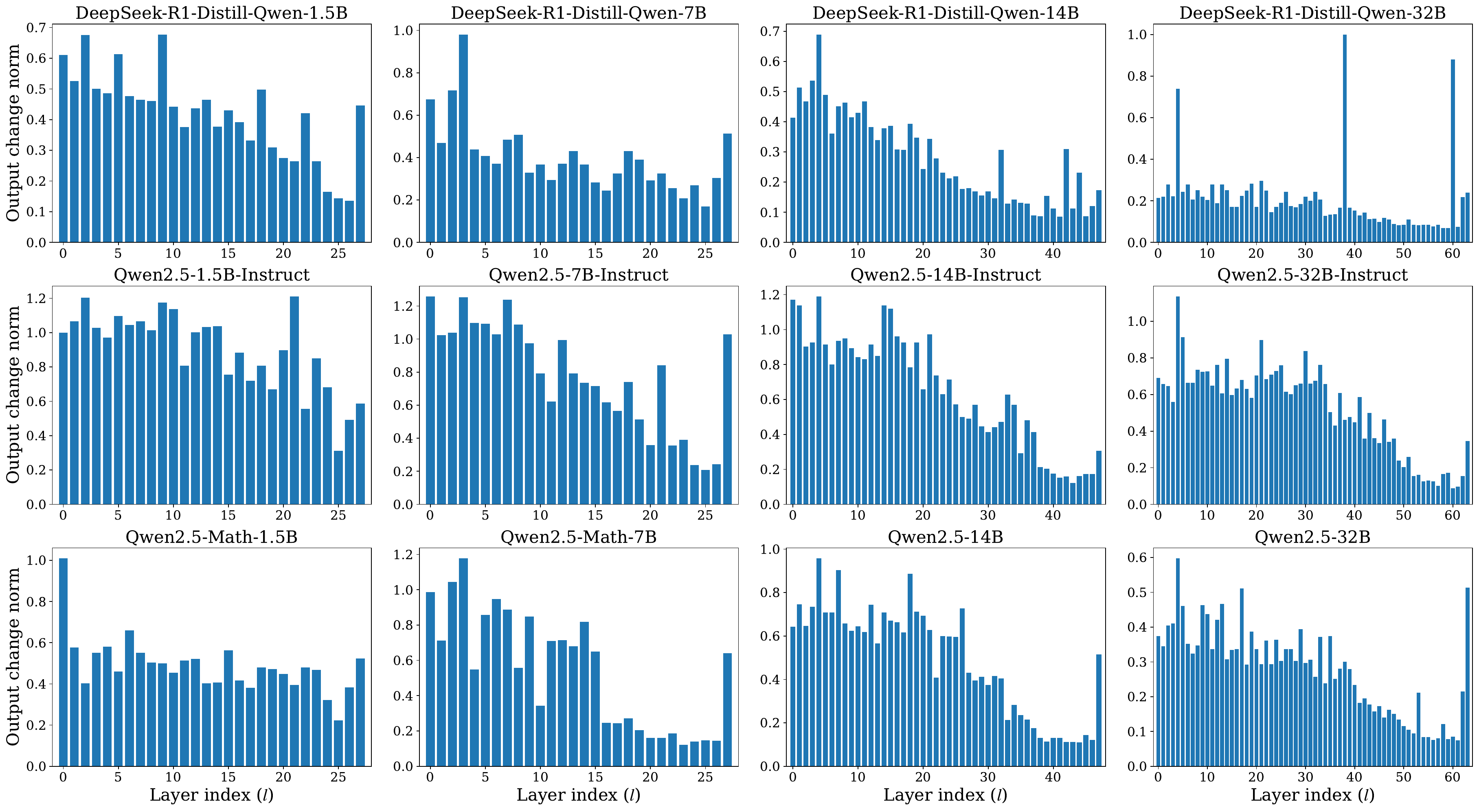}
        \label{fig:gsm8k_layer_effects_output_changes_all_models}
    }
    \caption{Effect of skipping a layer on future computation evaluated on GSM8K.}
    \label{fig:gsm8k_layer_effect}
\end{figure}

\begin{figure}[H]
    \centering
    \subfigure[Integrated gradients on addition.]{
        \includegraphics[width=.47\linewidth]{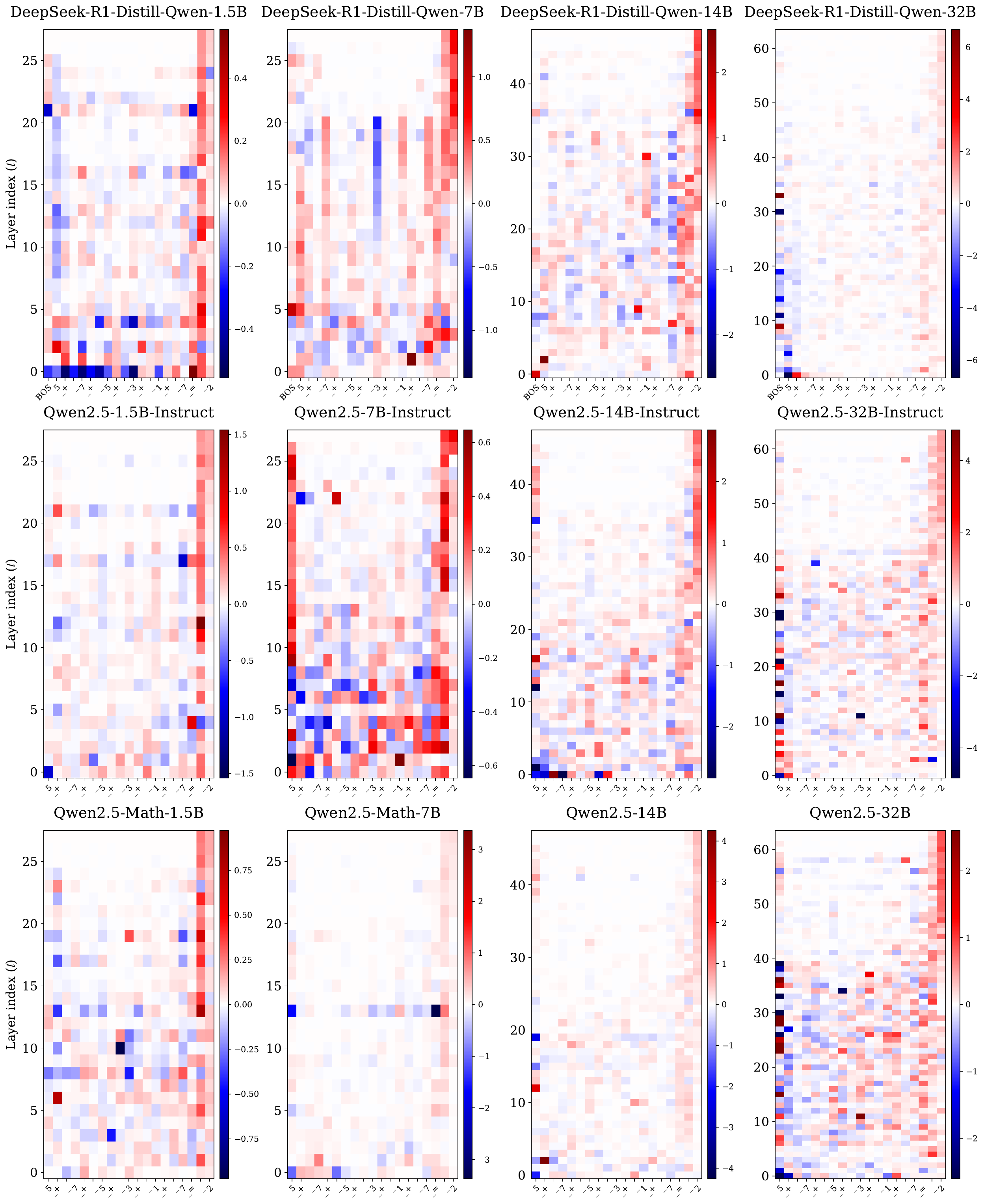}
        \label{fig:gsm8k_igrads_all_models}
    }
    \subfigure[Residual erasure on addition.]{
        \includegraphics[width=.47\linewidth]{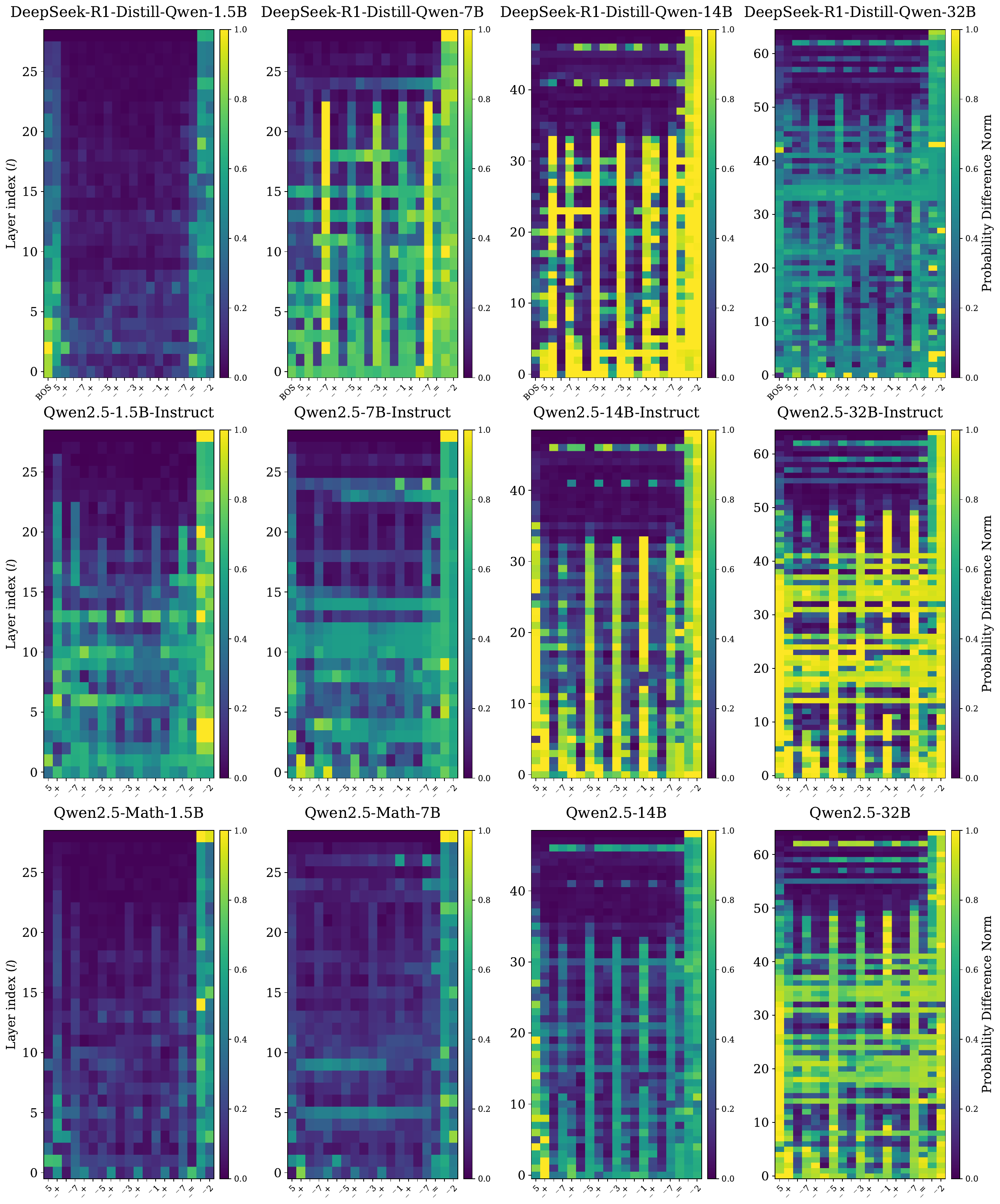}
        \label{fig:gsm8k_logit_erasure_all_models}
    }
    \caption{The Effects of individual computation steps evlauated on GSM8K.}
    \label{fig:gsm8k_individual_step}
\end{figure}

We show the results of logit lens in Figure~\ref{fig:gsm8k_logitlens}, the effects of skipping a layer on future computations in Figure~\ref{fig:gsm8k_layer_effects_all_models} and on output distributions in Figure~\ref{fig:gsm8k_layer_effects_output_changes_all_models}. Besides, the results of integrated gradients residual erasure are shown in Figure~\ref{fig:gsm8k_igrads_all_models} and Figure~\ref{fig:gsm8k_logit_erasure_all_models} respectively.

\section{Effective Depth of All Models Evaluated on GSM8K and HellaSwag}
\label{app:ed_of_gsm8k_and_hellaswag}
\begin{figure}[H]
    \centering
    \includegraphics[width=\linewidth]{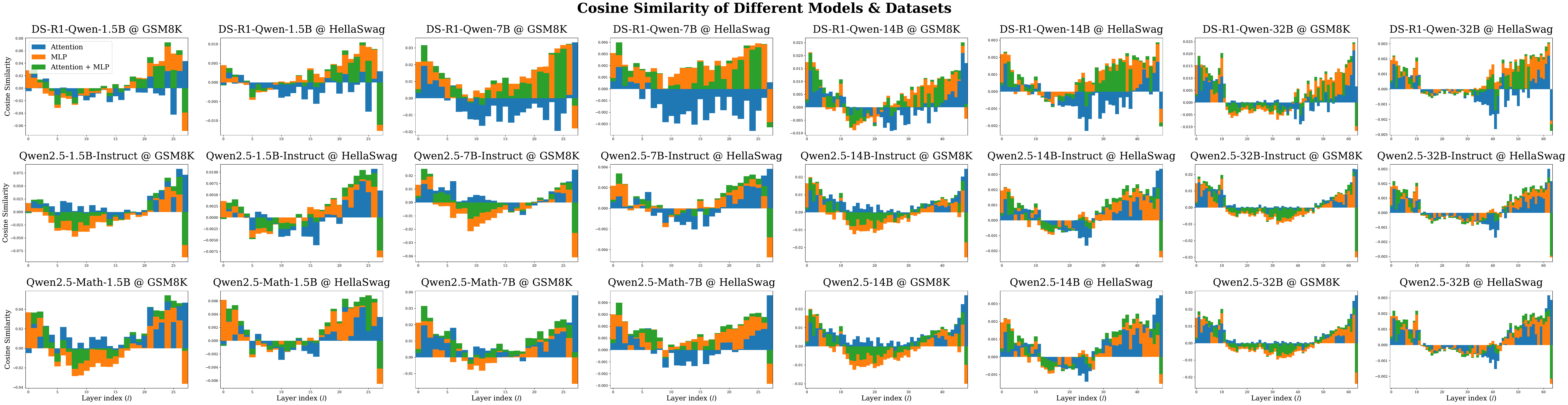}
    \caption{Residual cosine similarity of all models on GSM8K and HellaSwag.}
    \label{fig:gsm8k_hellaswag_cosine_similarity}
\end{figure}

\begin{figure}[H]
    \centering
    \includegraphics[width=\linewidth]{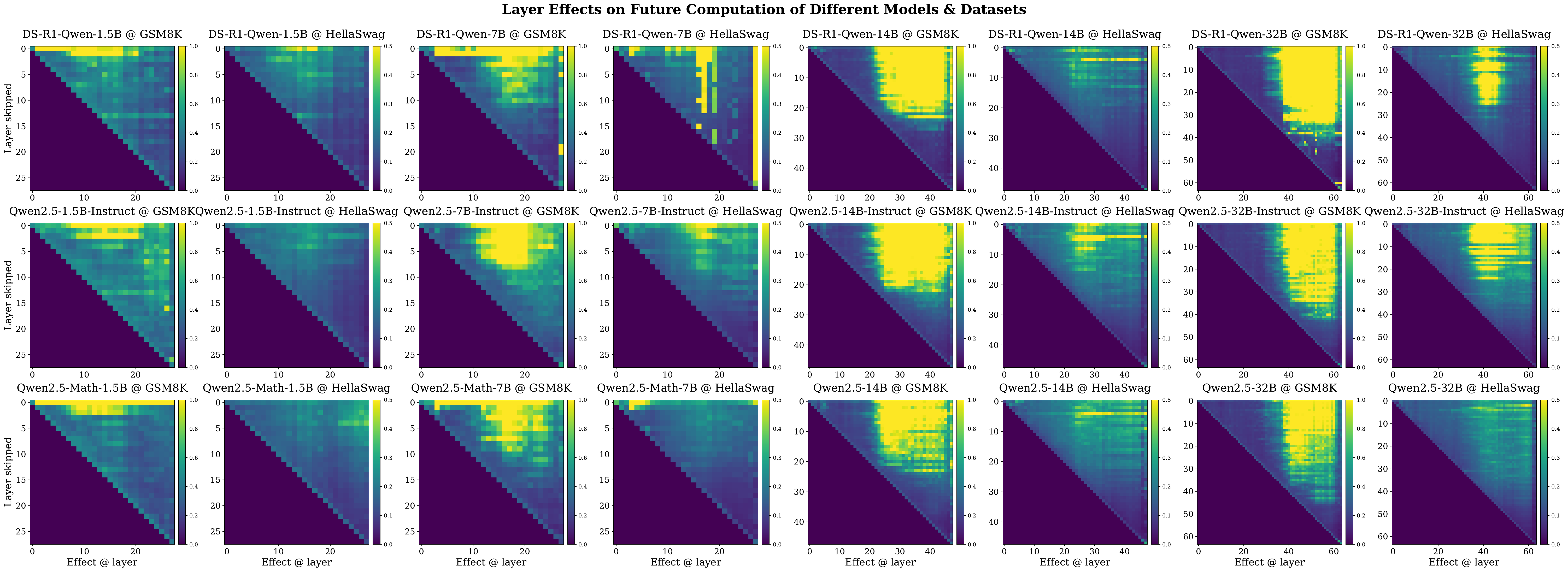}
    \caption{The effects of skipping a layer on future computations, the results include all models on GSM8K and HellaSwag.}
    \label{fig:gsm8k_hellaswag_layer_effects_future_compute}
\end{figure}

\begin{figure}[H]
    \centering
    \includegraphics[width=\linewidth]{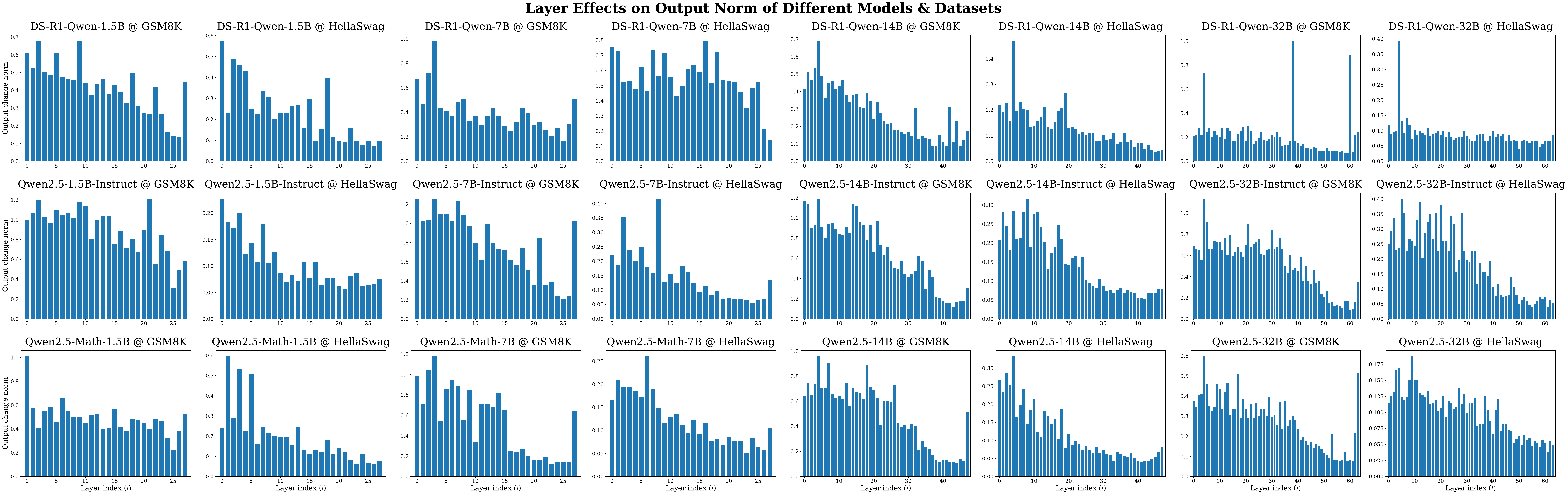}
    \caption{The effects of skipping a layer on output distributions, the results include all models on GSM8K and HellaSwag.}
    \label{fig:gsm8k_hellaswag_layer_effects_output_norm}
\end{figure}

\begin{figure}
    \centering
    \includegraphics[width=\linewidth]{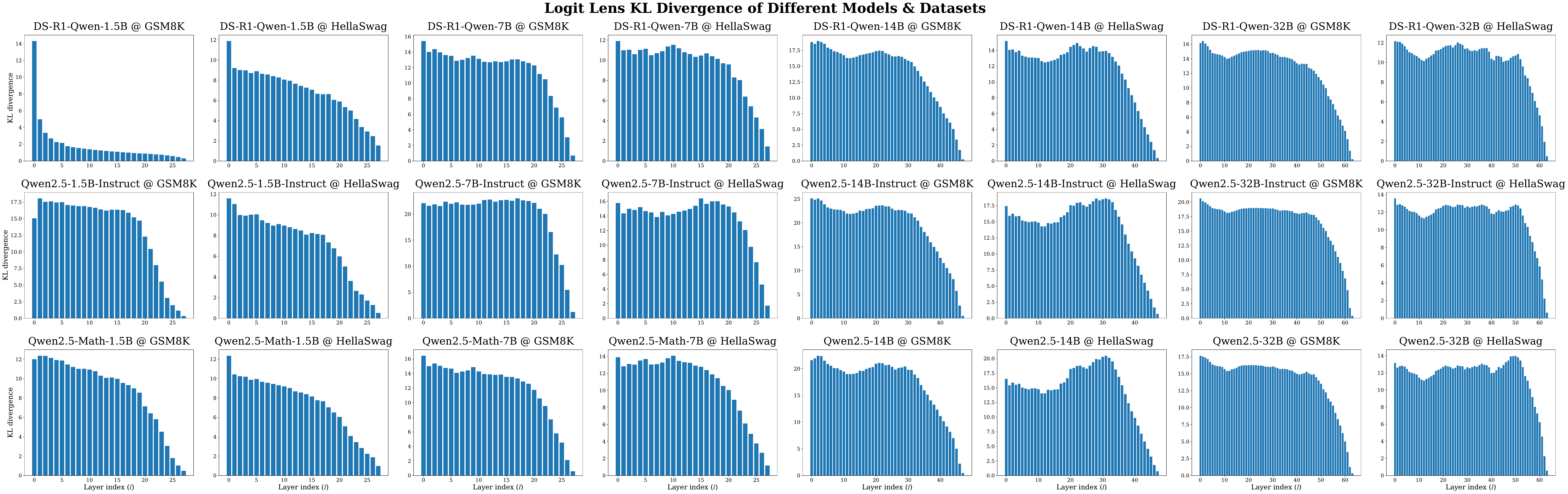}
    \caption{Logit lens KL divergence between early layer distributions and the final distributions. The results include all models on GSM8K and HellaSwag.}
    \label{fig:gsm8k_hellaswag_logit_lens_kl_divergence}
\end{figure}

\begin{figure}[H]
    \centering
    \includegraphics[width=\linewidth]{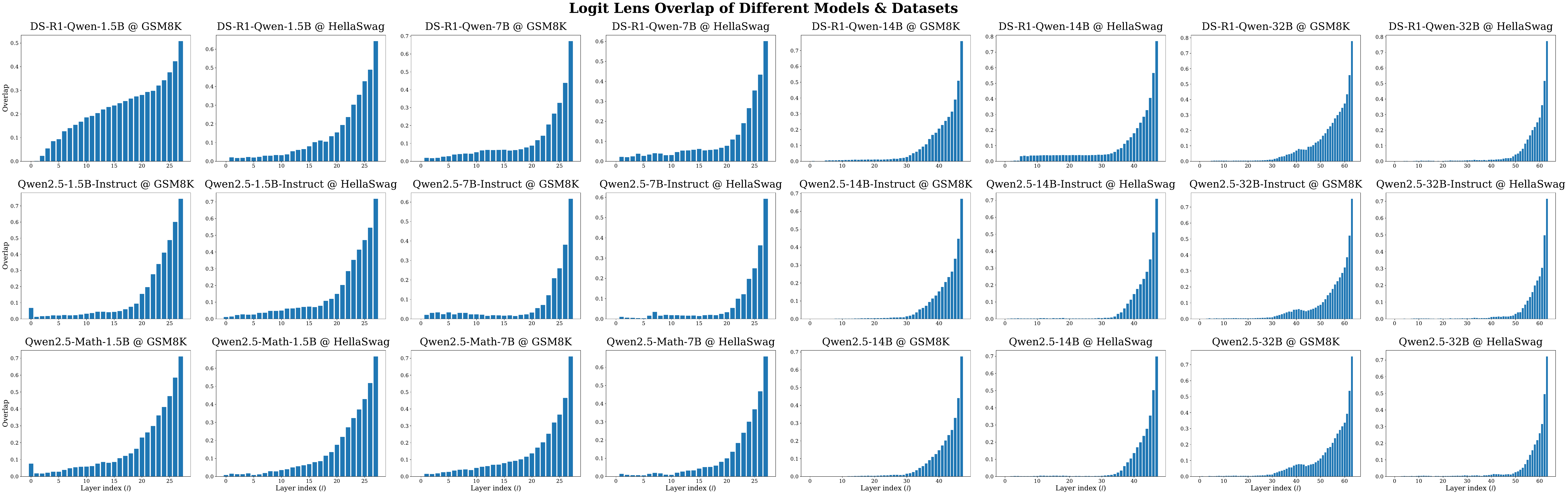}
    \caption{Logit lens top-5 overlap between early layer distributions and the final distributions. The results include all models on GSM8K and HellaSwag.}
    \label{fig:gsm8k_hellaswag_logit_lens_overlap}
\end{figure}

We show the results of effective depth of Qwen-2.5 family (base and instruct models) and their long-CoT variants tested on GSM8K and HellaSwag, including residual cosine similarity results in Figure~\ref{fig:gsm8k_hellaswag_cosine_similarity}; the effects of skipping a layer on future computations in Figure~\ref{fig:gsm8k_hellaswag_layer_effects_future_compute} and on output distributions in Figure~\ref{fig:gsm8k_hellaswag_layer_effects_output_norm}; logit lens KL divergence in Figure~\ref{fig:gsm8k_hellaswag_logit_lens_kl_divergence}; logit lens overlap in Figure~\ref{fig:gsm8k_hellaswag_logit_lens_overlap}.

\end{appendices}

\end{document}